\documentclass{article}

    \usepackage[final]{neurips_2025}

\usepackage[utf8]{inputenc} 
\usepackage[T1]{fontenc}    
\usepackage{hyperref}       
\usepackage{url}            
\usepackage{booktabs}       
\usepackage{amsfonts}       
\usepackage{nicefrac}       
\usepackage{microtype}      
\usepackage{xcolor}         

\usepackage{dsfont} 
\definecolor{bluejoint}{HTML}{0b0bff}  
\definecolor{orangejoint}{HTML}{fca200}  
\usepackage{pifont}
\usepackage{multirow}
\usepackage{bm}
\usepackage{graphicx}
\usepackage{wrapfig} 
\usepackage{bbm} 
\usepackage{amsmath}
\usepackage{algorithmic}
\usepackage{algorithm}
\usepackage{subcaption}  
\usepackage{natbib}
\setcitestyle{numbers,square}
\definecolor{myColor}{RGB}{26, 153, 230}
\newcommand{\bl}[1]{\textcolor{myColor}{#1}}
\usepackage[capitalize]{cleveref}
\crefname{section}{Sec.}{Secs.}
\Crefname{section}{Section}{Sections}
\Crefname{table}{Table}{Tables}
\crefname{table}{Tab.}{Tabs.}
\title{CSPCL: Category Semantic Prior Contrastive Learning for Deformable DETR-Based Prohibited Item Detectors}

\author{%
Mingyuan Li$^{1}$\quad Tong Jia$^{1,2}$\thanks{Corresponding author.}\quad Hao Wang$^{1}$\quad Bowen Ma$^{1}$\quad  \\
\textbf{Hui Lu$^{1}$\quad Shiyi Guo$^{1}$\quad Da Cai$^{1}$\quad  Dongyue Chen$^{1}$}\\
$^{1}$ College of Information Science and Engineering, Northeastern University, China\\
$^2$ State Key Laboratory of Synthetical Automation for Process Industries, Northeastern University\\
\texttt{542027743@qq.com}, \texttt{jiatong@ise.neu.edu.cn},  \texttt{ddsywh@yeah.net},\\ \texttt{ 2010285@stu.neu.edu.cn}, \texttt{2603813543@qq.com}, \texttt{guoshiyi@ise.neu.edu.cn}, \\\texttt{2210329@stu.neu.edu.cn}, \texttt{chendongyue@ise.neu.edu.cn}
}

\begin{document}

\maketitle

\begin{abstract}
Prohibited item detection based on X-ray images is one of the most effective security inspection methods. However, the foreground-background feature coupling caused by the overlapping phenomenon specific to X-ray images makes general detectors designed for natural images perform poorly. To address this issue, we propose a Category Semantic Prior Contrastive Learning (CSPCL) mechanism, which aligns the class prototypes perceived by the classifier with the content queries to correct and supplement the missing semantic information responsible for classification, thereby enhancing the model sensitivity to foreground features.
To achieve this alignment, we design a specific contrastive loss, CSP loss, which comprises the Intra-Class Truncated Attraction (ITA) loss and the Inter-Class Adaptive Repulsion (IAR) loss, and outperforms classic contrastive losses. Specifically, the ITA loss leverages class prototypes to attract intra-class content queries and preserves essential intra-class diversity via a gradient truncation function. 
The IAR loss employs class prototypes to adaptively repel inter-class content queries, with the repulsion strength scaled by prototype-prototype similarity, thereby improving inter-class discriminability, especially among similar categories.
CSPCL is general and can be easily integrated into Deformable DETR-based models. Extensive experiments on the PIXray, OPIXray, PIDray, and CLCXray datasets demonstrate that CSPCL significantly enhances the performance of various state-of-the-art models without increasing inference complexity.
The code is publicly available at \url{https://github.com/Limingyuan001/CSPCL}.

\end{abstract}
\section{Introduction}
\begin{wrapfigure}{l}{0.5\linewidth}
    \centering
    \includegraphics[width=1\linewidth]{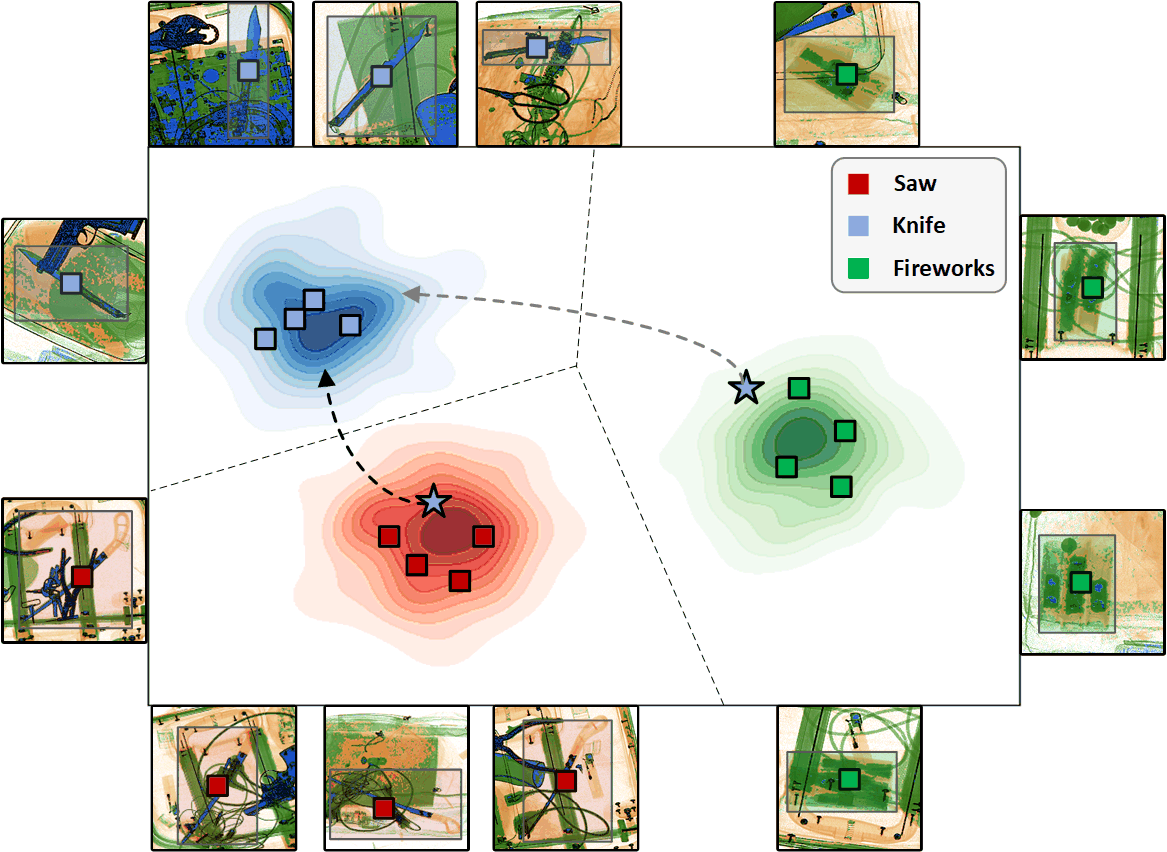}
    \caption{Feature manifold of saw (red), knife (blue), and fireworks (green). The squares and stars represent the feature representations of prohibited items and content queries, respectively.} 
    \vspace{-15pt} 
    \label{figure future manifold}
\end{wrapfigure}

Prohibited item detection based on X-ray images is an important security inspection measure, widely used in airports, postal services, government agencies, and border control areas. With the development of computer vision technologies, researchers have started using techniques such as image classification, object detection, and semantic segmentation to assist security personnel in examining packages imaged by security screening machines for prohibited items, thereby reducing the potential risks caused by work fatigue.
However, as shown in~\Cref{figure future manifold}, X-ray images differ from natural light images in that they exhibit unique overlapping phenomena, which cause coupling of foreground and background features. This leads to the detection results of general object detection models being easily affected by background noise.

There are two mainstream approaches for improving the anti-overlapping detection performance of detectors in response to the overlapping phenomenon in X-ray images. Methods based on attention mechanisms, including DAM~\cite{PIDray}, RIA~\cite{OPIXray}, and FCAM~\cite{MSDDet}, focus on foreground information either channel-wise or spatially. Methods based on label assignment, including IAA~\cite{GADet}, HSS~\cite{Xdet}, and LAreg\cite{CLCXray}, provide the model with more accurate candidate boxes during the training process, allowing it to concentrate on perceiving semantic information from overlapping foreground and background features.
However, the aforementioned methods are all designed for CNN-based detectors and lack a decoder structure to leverage the reliable prior knowledge provided by queries to perceive specific foreground features~\cite{DETR,DAB-DETR,DINO}.
Recently, researchers have found that clarifying the classification semantic information of content queries in Deformable DETR-based models can enhance the ability of the decoder to perceive foreground features in X-ray images and improve the anti-overlapping detection ability~\cite{AO-DETR,MMCL}.
AO-DETR~\cite{AO-DETR} introduces the CSA label assignment strategy, which constrains content queries to detect specific categories of objects, but its portability is relatively limited. 
MMCL~\cite{MMCL} proposes a plug-and-play contrastive learning mechanism that establishes sample pairs between content queries and categorizes them according to the number of classes.
This method indiscriminately repels inter-class sample pairs while attracting intra-class sample pairs, thereby clarifying the semantic information of categories. However, it lacks category prior information, which may lead to content queries not aligning with the true distribution of feature manifold of categories.

As shown in~\Cref{figure future manifold}, the features of prohibited items such as knife, saw, and fireworks exhibit distinct differences, with varying degrees of differentiation between categories. Specifically, knife and saw have more similar color, texture, and contour information compared to knife and fireworks. Consequently, the representations of knife and saw are closer together in feature space than those of knife and fireworks.
Based on this observation, we propose a novel Category Semantic Prior Contrastive Learning mechanism (CSPCL), which utilizes classifier weights as specific category prototypes to provide targeted correction and guidance to the content queries responsible for each category in the decoder layers, ensuring that the content queries align with the inherent characteristic distribution of the respective prohibited items.
To achieve this multi-class multi-sample alignment, we design a contrastive loss, called the Category Semantic Prior (CSP) loss, which comprises the Intra-Class Truncated Attraction (ITA) loss and the Inter-Class Adaptive Repulsion (IAR) loss, and shows superior performance compared to classic contrastive losses~\cite{N-pair, InfoNCE}.
Specifically, we employ the ITA loss to attract content queries of the same category toward their respective category prototypes, providing the queries with category semantic prior information.
It provides a gradient truncation function that stops the attraction when the similarity between the prototypes and intra-class content queries exceeds a certain threshold, thereby preventing the homogenization of intra-class content queries.
Inversely, the IAR loss utilizes class prototypes to repel inter-class content queries, which has a repulsion factor adaptively adjusting the repulsion strength based on the similarity between prototype pairs, thereby facilitating inter-class content queries to learn discriminative identifying features of similar categories, and obtain adequate separability.


The main contributions of our work are as follows:
    \begin{itemize}
    \item We propose a plug-and-play CSPCL mechanism that uses contrastive learning to align content queries with the intrinsic feature distribution of the same category of prohibited items as perceived by the classifier, thereby enhancing the anti-overlapping detection capability of Deformable DETR-based models without increasing inference complexity.
    \item We propose the CSP loss tailored for the multi-class multi-sample alignment task, which more effectively aligns class prototypes with content queries than classic contrastive losses.
    \item The CSPCL mechanism demonstrates strong generalization across various Deformable DETR variants, such as RT-DETR~\cite{RT-DETR}, DINO~\cite{DINO}, and AO-DETR~\cite{AO-DETR}, improving detection accuracy on two prohibited item datasets, including PIXray~\cite{PIXray} and OPIXray~\cite{OPIXray}.
    Furthermore, we validate CSPCL on the large-scale prohibited item detection datasets PIDray~\cite{PIDray} and CLCXray~\cite{CLCXray}, further advancing the state-of-the-art standards.
\end{itemize}

\section{Related Work}
\textbf{DETR-like Detectors.}
DEtection TRansformer (DETR) pioneered the removal of the Non-Maximum Suppression (NMS) post-processing step, streamlining the pipeline of object detectors. However, it still has two major drawbacks: slow training convergence and the ambiguous meaning of queries~\cite{DETR}. 
To address the issue of slow convergence, Deformable DETR~\cite{Deformable-DETR} introduces deformable attention, which only attends to a small set of key sampling points around one selected reliable reference point, significantly improving the training convergence speed. Stale-DINO~\cite{Stable-DETR} uses the IoU score as ground truth to supervise the predicted classification scores of positive examples, thereby improving the stability of matching. Group-DETR~\cite{Group-DETR} proposes a group-wise one-to-many assignment that conducts one-to-one assignment within each group of object queries, resulting in one ground-truth object being assigned to multiple predictions. H-DETR~\cite{H-DETR} introduces an auxiliary one-to-many matching branch during training. Co-DETR~\cite{CO-DETR} integrates multiple traditional one-to-many matching strategies.
To clarify the meaning of queries, Conditional DETR~\cite{Conditional-detr} introduces 2D reference points in the query part as reliable spatial information hints. Anchor-DETR~\cite{Anchor-DETR} and DAB-DETR~\cite{DAB-DETR} introduce 4D reference boxes to further provide potential position and size information of the objects. DN-DETR~\cite{DN-DETR} proposes denoising training to help the network select high-quality queries, whose spatial positions are closer to the ground truth, to predict bounding boxes. DINO~\cite{DINO} further proposes contrastive denoising training, teaching the network how to reject queries that are far from the ground truth. 

\textbf{Contrastive Learning.}
The core idea of contrastive learning is to train more discriminative feature representations by pulling together the representations of similar samples and pushing apart the representations of dissimilar samples. Overall, contrastive learning can be divided into unsupervised contrastive learning and supervised contrastive learning.
Unsupervised contrastive learning methods can be further divided into instance-wise contrastive learning and cluster-based contrastive learning. The former is represented by methods such as InstDisc~\cite{InstDisc} and SimCLR~\cite{SimCLR}, where, for any given instance, its data-augmented version is treated as an intra-class sample, while other instances are considered as inter-class samples. Typically, the InfoNCE~\cite{InfoNCE} loss is used for gradient updates. The latter, represented by methods such as CC~\cite{CC} and TCL~\cite{TCL}, uses pseudo-labels generated by clustering algorithms as the basis and then employs a supervised contrastive learning framework for training.
Supervised contrastive learning methods have three paradigms of loss function. The first is based on the LMMN~\cite{LMMN}, which includes Triplet loss and N-pair loss. The second is based on the Softmax function, such as AM-Softmax~\cite{AM-Softmax}, Circle loss~\cite{Circle_loss}, and SupCon loss~\cite{SupCon_loss}. The third paradigm is based on Cross-Entropy loss, including methods like EBM~\cite{EBM}, C2AM~\cite{C2AM}, and MMCL~\cite{MMCL}. As far as we know, the choice among these three paradigms often depends on the specific task and the reliability of the labels, with no definitive superiority among them.
In this paper, we propose a targeted contrastive loss function, CSP loss, for the task of guiding content queries with category semantic priors.

\textbf{Prohibited Item Detector.}
Since the introduction of the first large-scale pseudo-color X-ray prohibited item detection dataset by SIXray~\cite{SIXray}, research on prohibited item object detection in X-ray images has become increasingly diverse and manifold. Prohibited item detectors are typically based on general object detectors and are improved to address the overlap phenomena in X-ray images~\cite{tao2025dual,DvXray,tao2022few,GADet,tao2022exploring,ForkNet,tao2021towards}. SIXray introduced the CHR mechanism to iteratively strip away overlapping background information. OPIXray~\cite{OPIXray} proposed the DOAM module to perceive the material and edge information of foreground objects. GADet~\cite{GADet} and Xdet~\cite{Xdet} proposed the IAA and HSS label assignment strategies to mitigate the foreground-background class imbalance issue caused by overlapping phenomena. FDTNet~\cite{FDTNet} and FAPID~\cite{FAPID} transform features into the frequency domain and then use self-attention mechanisms or convolutions to extract the texture and edge information of foreground objects. AO-DETR~\cite{AO-DETR} proposed the first DETR-like model in the field of prohibited item detection, utilizing the CSA strategy to train category-specific content queries that are responsible for detecting specific types of contraband, thereby enhancing the ability to detect overlapping objects. However, the portability of CSA is limited.
MMCL~\cite{MMCL} proposed a plug-and-play contrastive learning strategy that helps clarify the category-specific semantic information of content queries in Deformable DETR-based models. However, it lacks prior knowledge of the feature distribution for each category, and the indiscriminate repulsion of inter-class content queries will lead to the misalignment between the feature distributions of queries and objects within the same category. This paper presents improvements to address this issue.
\begin{figure*}
    \centering
    \includegraphics[width=1\linewidth]{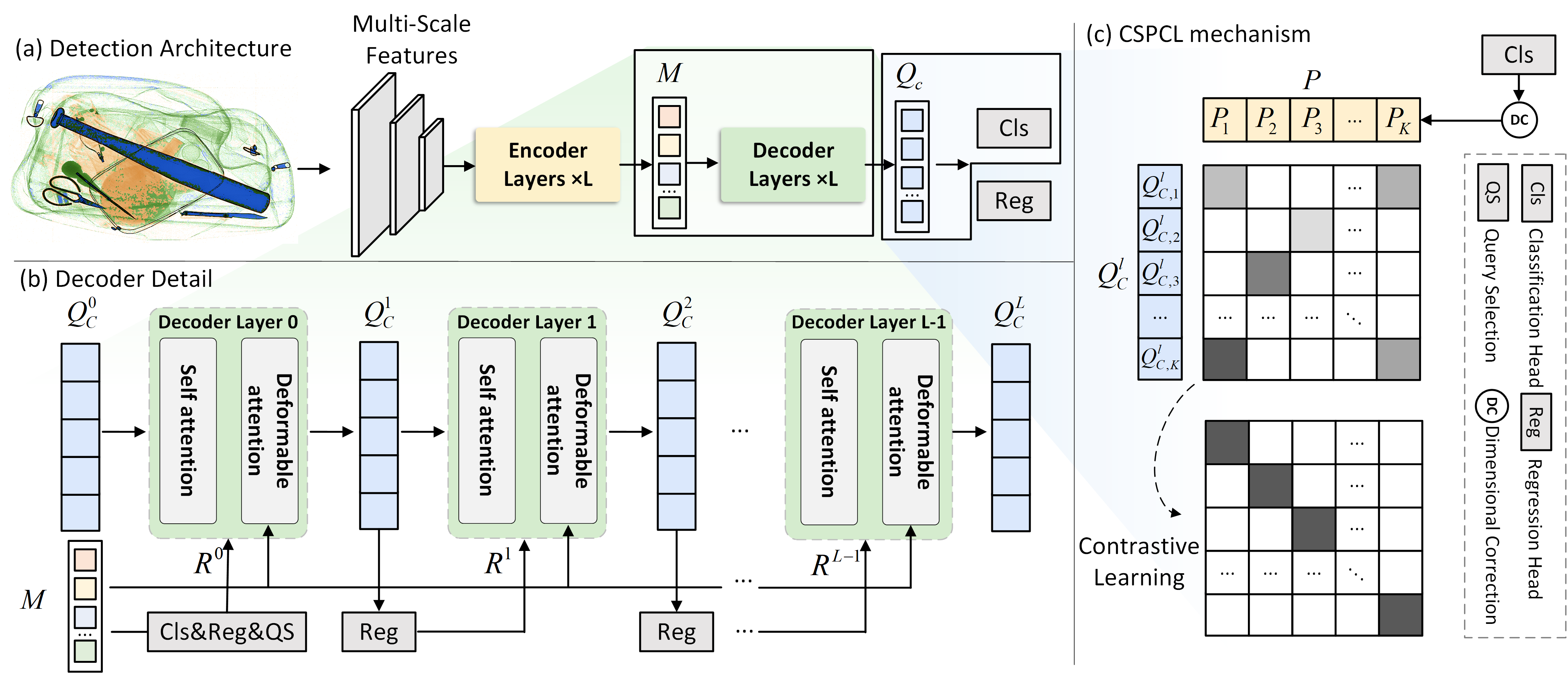}
    \caption{
Illustrating the pipeline of CSPCL plugged into one Deformable DETR~\cite{Deformable-DETR} variant DINO~\cite{DINO}.
(a) The overall architecture of DINO.
(b) The process of updating content queries in each layer of the decoder.
(c) The CSPCL mechanism uses contrastive loss to leverage classifier weights as class prototypes, thereby supplementing and refining the semantic information responsible for classification in the content queries of the decoder at layer $l$.
    }
    \vspace{-10pt} 
    \label{figure overall architecture1}
\end{figure*}
\section{Proposed Method}
\subsection{Explore the significance of feature alignment between objects and content queries}\label{subsection explore}
In this part, we examine the role of content queries in the decoder of Deformable DETR-based models and elucidate the importance of aligning object features with content queries.
As shown in~\Cref{figure overall architecture1}(a), using DINO as a representative model of the Deformable DETR series, given an input image, the backbone of the model first extracts multi-scale features, which are then integrated through multiple layers of encoder layers to obtain global information, commonly referred to as the encoder memory $\mathbf{M}\in \mathds{R}^{N\times C}$~\cite{DETR}, which is passed to the decoder. In the decoder, as depicted in~\Cref{figure overall architecture1}(b), the decoder first uses the $\mathbf{M}$ to predict a set of candidate results, including classification results $\mathbf{C}\in \mathds{R}^{N\times K}$ and localization results $\mathbf{R}\in \mathds{R}^{N\times 4}$, where $K$ is the number of categories in the dataset. Then, through the query selection mechanism, the top $N_{pred}$ most reliable predictions $\mathbf{R}^0\in \mathds{R}^{N_{pred}\times 4}$ are filtered based on their classification confidence.

Overall, the decoder refines and updates the randomly initialized content queries $\mathbf{Q}_c^0$ through $L$ layers of decoder layers, using the encoder memory $\mathbf{M}$ and spatial prior information $\mathbf{R}^0$. This iterative process yields more accurate classification and localization results. The decoder layer at the $l$-th layer $\mathcal{D}^l$, along with the corresponding prediction head, can be expressed by the following formulas:
{\small
\begin{equation}
\label{equation decoder}
\{\mathbf{Q}_c^{l+1},\mathbf{R}^{l+1},\mathbf{C}^{l+1}\}=\mathcal{D}^l(\mathbf{Q}_c^l,\mathbf{R}^l,\mathbf{M};\theta^l),\quad
\mathbf{R}^{l+1}=\sigma(\sigma^{-1}(\mathbf{R}^l)+\Delta \mathbf{R}^{l+1}),
\end{equation}
\begin{equation}
\label{equation FFN 2}
\{\Delta \mathbf{R}^{l+1}, \mathbf{C}^{l+1}\}=\mathcal{H}_{R,C}^l(\mathbf{Q}_c^{l+1};\theta^l_\mathcal{H}),
\end{equation}
}
where $l\in \{x \mid x\in \mathds{Z}, 0\leq x < L\}$ is the decoder block index, and $L$, topically set to 6, denotes the total number of decoder blocks. 
The $\theta$, $\sigma(\cdot)$, $\sigma^{-1}(\cdot)$, and $\mathcal{H}_{R,C}$ denote learnable parameters, sigmoid, inverse sigmoid~\cite{Deformable-DETR}, and prediction heads of regression and classification~\cite{DETR}, respectively. 
\cref{equation FFN 2} shows that content queries almost directly determine the prediction results at each layer of the decoder. To analyze the update process of content queries more deeply, we further decompose the decoder in~\cref{equation decoder} into self-attention mechanisms and deformable attention mechanisms for separate explanations.

The self-attention $\mathcal{D}_s^l$ first performs element-wise addition to fuse the spatial information $\mathbf{Q}_p^l$, which is obtained by positional embedding of reference boxes $\mathbf{R}^l$~\cite{DINO}, with the content query $\mathbf{Q}_c^l$, resulting in the query of self-attention $\mathbf{Q}_s^l$ and the key of self-attention $\mathbf{K}_s^l$.
This establishes a global mapping, and by extracting and aggregating features in $\mathbf{V}_s^l$ that have strong correlations with the spatial information, the output $\mathbf{Q}_{s,c}^l$ is obtained. The $l$-th self-attention layer can be denoted as follows:
{\small
\begin{equation}
\label{equation self 1}
\mathbf{Q}_{s,c}^l=\mathcal{D}_s^l(\mathbf{Q}_s^l,\mathbf{K}_s^l,\mathbf{V}_s^l;\theta_s^l),\quad
\mathbf{Q}_s^l=\mathbf{K}_s^l=\mathbf{Q}_c^l+\mathbf{Q}_p^l,\quad
\mathbf{V}_s^l=\mathbf{Q}_c^l,
\end{equation}
}
where $\mathbf{K}$ and $\mathbf{V}$ are the key and value~\cite{Attention_is_all_you_need}, respectively. 

Deformable attention $\mathcal{D}_d^l$ uses reference boxes $\mathbf{R}^l$ to assist the deformable queries $\mathbf{Q}_d^l$, element-wise addition result of self-attention output $\mathbf{Q}_{s,c}^l$ and spatial embeddings $\mathbf{Q}_p^l$, and extract feature information from the encoder memory $\mathbf{M}$ to update the content queries $\mathbf{Q}_c^{l+1}$.
The $l$-th deformable attention layer can be denoted as follows:
{\small
\begin{equation}
\label{equation deform 1}
\mathbf{Q}_c^{l+1}=\mathbf{Q}_{d,c}^{l+1}=\mathcal{D}_d^l(\mathbf{Q}_d^l,\mathbf{R}^l,\mathbf{M};\theta_d^l),\quad
\mathbf{Q}_d^l=\mathbf{Q}_{s,c}^l+\mathbf{Q}_p^l.
\end{equation}
}
From the above process, it can be seen that content queries $\mathbf{Q}_c^l$ are continuously integrated with positional information from $\mathbf{Q}_p^l$ through self-attention and deformable attention mechanisms.
As the number of iterations in the decoder layer increases, the inherent category semantics in the original content query are gradually updated and replaced by the $\mathbf{Q}_p^l$ that contains location information such as contours, edges, and shapes. This results in the loss of the content query’s role in providing classification-related semantic information, such as color and texture. 
Therefore, if we can supplement and correct the category semantics in the content queries, we can enhance the model's sensitivity to the corresponding foreground object features.

\subsection{Category semantic prior contrastive learning (CSPCL)}
We propose the CSPCL mechanism, as illustrated in~\Cref{figure overall architecture1}(c), which supplements and corrects the category semantic information of content queries by aligning them with the class prototypes perceived by the classifier weights.
In the preprocessing stage, we group the content queries $\mathbf{Q}_c^l\in \mathds{R}^{N_{pred}\times M}$ from the $l$-th layer based on the number of categories $K$, resulting in $K$ groups of category-specific content queries $\mathbf{Q}_{c,k}^l\in \mathds{R}^{n\times M}$, where $k\in \{x \mid x\in \mathds{Z}, 0< x \leq  K\}$, and $n$ is the quotient of the number of content queries $N_{pred}$ divided by the number of categories $K$. Next, the classifier weights $\mathbf{W}\in \mathds{R}^{K\times M}$ are dimensionally expanded and adjusted to align with the dimensions of $\mathbf{Q}_c^l$, resulting in prototypes $\mathbf{P}\in \mathds{R}^{N_{pred}\times M}$. 
In this way, the prototype, $\mathbf{P}_k\in \mathds{R}^{n\times M}$, which is responsible for category $k$, shares the same dimensions as $\mathbf{Q}_{c,k}^l$. 
The prototypes $\mathbf{P}_k$ and the content queries $\mathbf{Q}_{c,k}^l$ consist of $n$ sample pairs for each category $k$. For this multi-class multi-sample alignment task, we specifically designed the CSP loss consisting of the Intra-Class Truncated Attraction (ITA) loss and the Inter-Class Adaptive Repulsion (IAR) loss. 
The IAR loss employs category prototypes to repel inter-class content queries, while the ITA loss attracts content queries towards the corresponding intra-class prototypes, thereby enhancing inter-class discriminability---especially among similar categories---while preserving essential intra-class diversity.
The formula for CSP loss is as follows:
{\small
\begin{equation}
\label{equation CSPCL}
\mathcal{L}_{CSP}(\mathbf{P,Q},K)=\alpha \mathcal{L}_{ITA}(\mathbf{P,Q},K)+\beta \mathcal{L}_{IAR}(\mathbf{P,Q},K),
\end{equation}
}
wherein $\alpha$ and $\beta$ are hyperparameters used to balance two losses.

\begin{figure}[h]
    \centering
    \includegraphics[width=1\linewidth]{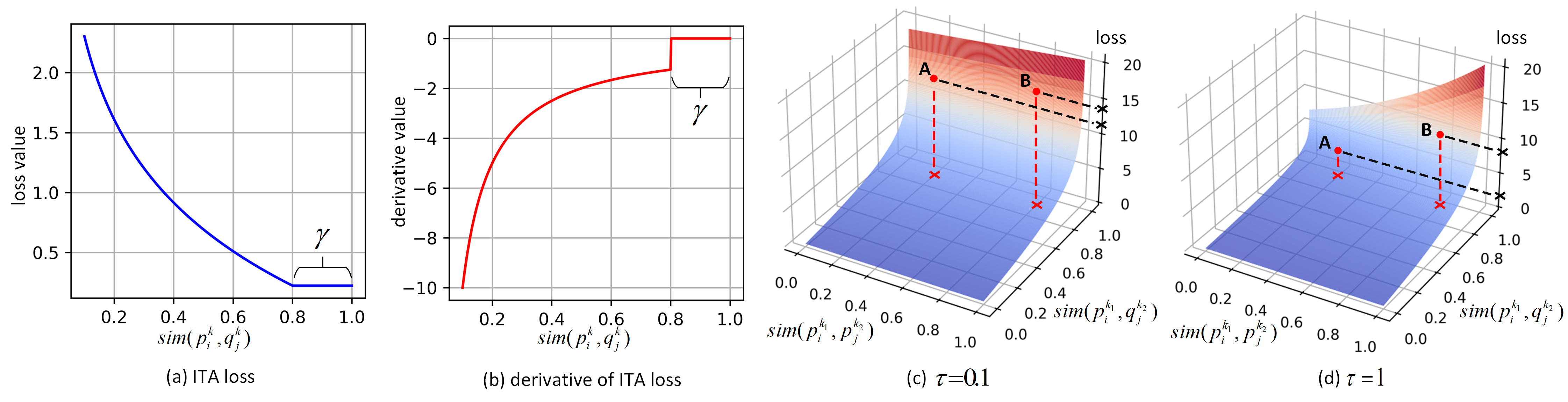}
    \caption{(a) and (b) are the curves of the ITA loss and its derivative, respectively. (c) and (d) are the 3D surface plots of our IAR loss with different $\tau$.
    The samples A and B have the same $\text{sim}(\mathbf{p}_i^{k_1},\mathbf{q}_j^{k_2})$, but $\text{sim}(\mathbf{p}_i^{k_1},\mathbf{p}_j^{k_2})$ of A is smaller that of B. When $\tau=0.1$, the IAR loss values of A and B are almost the same, but when $\tau=1$, The loss value of A is much smaller than that of B.}
    \label{figure intra class loss}
\end{figure}

\textbf{ITA loss.} To attract category-specific content queries to the feature manifold of intra-class prohibited items, a simple and effective method is to compute the cross-entropy loss using the cosine similarity between category prototypes and intra-class content queries. However, even prohibited items of the same category still exhibit feature variations under different backgrounds and poses, meaning that intra-class content queries need to maintain a certain degree of variability to align with these differences, thereby improving the detection accuracy. Therefore, we specifically designed a gradient truncation mechanism to ensure that content queries retain enough variance. The overall formula for the ITA loss is as follows:
{\footnotesize
\begin{equation}
\label{equation intra}
\mathcal{L}_{ITA}(\mathbf{P,Q},K)=\frac{-1}{Kn(n-1)}\sum_{k=1}^K\sum_{i=1}^n\sum_{j=1}^n \text{log}\Big(\mathcal{T}\big(\text{sim}(\mathbf{p}_i^k,\mathbf{q}_j^k),\gamma\big)\Big),
\mathcal{T}(x,\gamma)=
\begin{cases}
1-\gamma,  & 1-\gamma < x\le 1\\
x, & 0< x < 1-\gamma \\
0, &-1\le x < 0
\end{cases}    
\end{equation}
}
where $\text{sim}(\mathbf{p}_i^k,\mathbf{q}_j^k)$ is the cosine similarity between the $i$-th prototype of the $k$-th class and $j$-th content query of the $k$-th class. 
$\mathcal{T}(x,\gamma)$ is the gradient truncation function. As shown in~\Cref{figure intra class loss}, when the input value $x$, i.e. $\text{sim}(\mathbf{p}_i^k,\mathbf{q}_j^k)$, exceeds the threshold $1-\gamma$, the returned ITA loss value is constant $1-\gamma$, and its derivative becomes $0$. At this point, the prototype stops attracting the intra-class content queries, allowing the content queries to retain the necessary variability and margin among queries. Therefore, by adjusting the value of $\gamma$, we can control the variance (diversity) of similarities of intra-class content queries. 


\textbf{IAR loss.}
Although the ITA attracts content queries toward the intra-class prototypes, some stubborn samples still drift away from the sample center, and may even appear in the feature manifold of other categories. A simple solution is to use the prototype to repel inter-class content queries indiscriminately.
However, the differences between the prototypes of different categories vary naturally. As shown in~\Cref{figure future manifold}, the feature manifolds of knife and saw have less margin than those of knife and firework, which means the former is naturally more coupled than the latter, requiring stronger repulsion by the prototype to create enough margin for clarifying the semantic information of content queries for better detection performance~\cite{LDAM,Theorem1ofLDAM,AO-DETR,MMCL}. Therefore, we propose the IAR loss as follows:
\vspace{-10pt} 
\begin{equation}
\label{equation inter}
\mathcal{L}_{IAR}(\mathbf{P,Q},K)=\frac{-1}{K(K-1)n^2}\sum_{k_1=1}^K\sum_{k_2=1}^K\sum_{i=1}^n\sum_{j=1}^n \mathds{1}\left[ k_1\neq k_2\right] \\
\mathcal{R}(k_1,k_2)\text{log}(1-\text{sim}(\mathbf{p}_i^{k_1},\mathbf{q}_j^{k_2})),
\end{equation}
\begin{equation}
\label{equation margin}
\mathcal{R}(k_1,k_2)=e^{1-\tau \cdot\big(1-\text{sim}(\mathbf{p}_i^{k_1},\mathbf{p}_j^{k_2})\big)},
\end{equation}

where $\mathds{1}[ k_1\neq k_2] \in \{0,1\}$ is an indicator function. It equals $1$ if $k_1\neq k_2$, and $0$ in the other case. Additionally, $\mathcal{R}(k_1,k_2)$ is the repulsion factor, used to adjust the strength of the repulsive effect exerted by the category prototype on content queries from different categories. The higher the similarity between the prototype of the $k_1$-th category $\mathbf{p}_i^{k_1}$ and the prototype of the $k_2$-th category $\mathbf{p}_j^{k_2}$, the stronger the repulsion exerted by the prototype $\mathbf{p}_i^{k_1}$ on the content query $\mathbf{q}_j^{k_2}$.
$\tau$ is the temperature coefficient, which controls the sensitivity of the IAR loss to inter-class prototype similarity. As shown in~\Cref{figure intra class loss}, the larger the $\tau$, the more sensitive the IAR loss becomes to differences between prototypes. This means that the prototype only repels content queries belonging to similar categories while minimizing the influence on content queries from unrelated categories, which helps align the features between prototypes and queries.
When $\tau=0$, $\mathcal{R}(k_1,k_2)$ becomes a constant, and the prototypes indiscriminately repel content queries from all categories.
Therefore, inter-class content queries can learn discriminative identifying features for similar categories, through our IAR loss with a satisfied $\tau$.

\subsection{Plug CSPCL into target decoder layers}
The CSPCL mechanism can be inserted into the decoder of Deformable DETR-like models in a plug-and-play manner without increasing model complexity, as described in~\cref{Algorithm CSPCL}.
Given the number of classes $K$ in the dataset, the content queries $\mathbf{Q}_c$, localization results set $R$ and classification results set $C$ of all decoder layers set $L$, the class prototypes $\mathbf{P}$, and the set of target layers $\hat{L}$ where our CSPCL mechanism will be inserted.
For each decoder layer $l$, the model's classification and localization detection results are matched with the ground truth through the Hungarian matching mechanism~\cite{DETR} to establish a one-to-one correspondence. Then, we compute the built-in loss $\mathcal{L}_{BASE}$ of the model for the current layer $l$ based on the matched prediction result and ground truth pairs.

Additionally, we determine whether the current layer belongs to the target layers set $\hat{L}$. If it does, we use the content queries from the current layer $\mathbf{Q}_c^l$, prototypes $\mathbf{P}$, and the number of categories $K$ to compute the CSP loss according to~\cref{equation CSPCL}. This loss is then added to the model's inherent loss to update the current layer's loss value $\mathcal{L}^l$. Finally, we use the sum of the losses from each layer, denoted as $\mathcal{L}$, to update both the model parameters and the content queries of all layers.
\begin{wrapfigure}{r}{0.5\linewidth}
\vspace{30pt} 
\begin{minipage}{\linewidth} 
\begin{algorithm}[H]
\caption{Plug CSPCL into the Decoder.}\label{Algorithm CSPCL}
\begin{algorithmic}
\REQUIRE ~~\\ 
Constant $K$;\quad
Set $L$, $\hat{L}$, $\mathbf{Q}_c$, $\mathbf{P}$, $\mathbf{R}$, $\mathbf{C}$, $\mathbf{G}$;\
\ENSURE ~~\\ 
\STATE Initialize the total loss $\mathcal{L}$ to 0;
\FOR{ $\forall$ decoder layer index $ l\in L$ }
\STATE $\{ \mathbf{R}_i^l, \mathbf{C}_i^l; \mathbf{G}_i \}\gets \mathcal{H}^l(\mathbf{R}^l,\mathbf{C}^l;\mathbf{G})$;
\STATE $\mathcal{L}^l \gets$ $\mathcal{L}_{BASE}(\{\mathbf{R}_i^l, \mathbf{C}_i^l; \mathbf{G}_i \})$;
\IF{$l \in \hat{L}$}
\STATE $\mathcal{L}^l \gets \mathcal{L}^l + \mathcal{L}_{CSP}(\mathbf{P}, \mathbf{Q}_c^l, K)$;
\ENDIF
\STATE $\mathcal{L} \gets \mathcal{L}+\mathcal{L}^l$;
\ENDFOR
\STATE update model parameters and $\mathbf{Q}_c$ to minimize $\mathcal{L}$;
\RETURN $\mathbf{Q}_c$;
\end{algorithmic}
\label{algorithm1}
\end{algorithm}
\end{minipage}
\vspace{-35pt} 
\end{wrapfigure}
\subsection{Effect of the CSPCL mechanism}

\begin{figure}[h]
    \centering
    \includegraphics[width=1\linewidth]{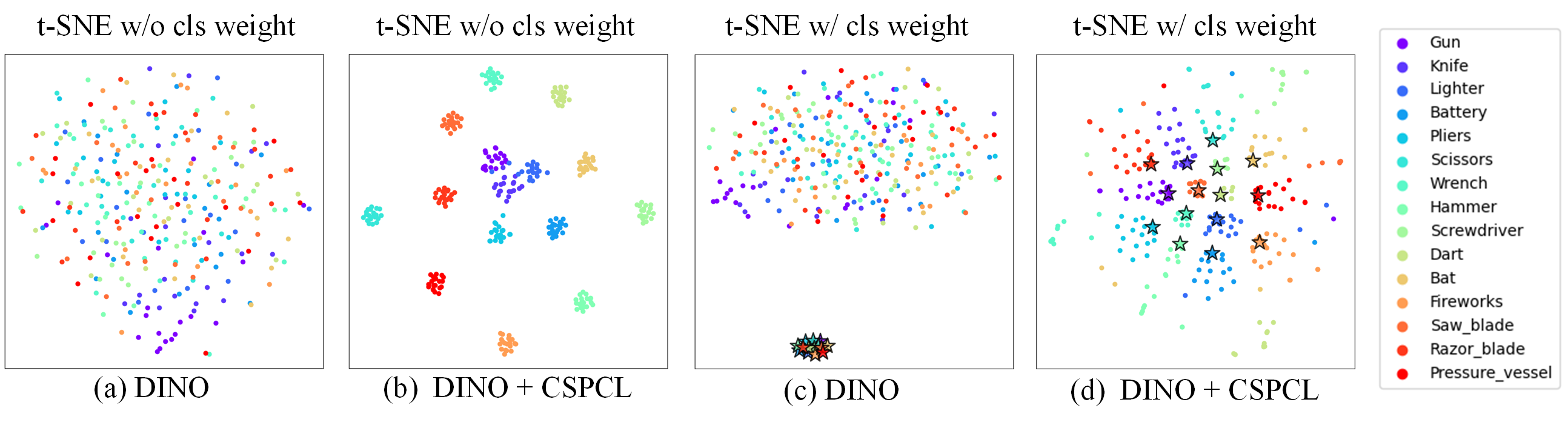}
    \caption{
    The t-SNE visualization results of the content queries (dots) and classifier weights (stars) from the first decoder layer. (a) and (b) show the visualization of content queries for DINO and DINO with the CSPCL mechanism, respectively. (c) and (d) show the visualization results of the classifier weights and content queries simultaneously reduced for the two models, DINO and DINO with the CSPCL mechanism, respectively.
    }
    \label{figure t-SNE 2}
\end{figure}

To analyze the effectiveness of the CSPCL mechanism for content queries, we visualize their t-SNE~\cite{t-SNE} dimensionality reduction distribution results with classifier weights.
As shown in~\Cref{figure t-SNE 2}(a) and~\Cref{figure t-SNE 2}(c), content queries of all categories of the original DINO are scattered and evenly distributed in the feature space, while also exhibiting significant differences from the category prototypes perceived by the classifier.
This suggests that during the model training process, the content queries in the decoder fail to capture the semantic information responsible for classification, which is consistent with the conclusion drawn in~\Cref{subsection explore}. 
In contrast, \Cref{figure t-SNE 2}(b) and \Cref{figure t-SNE 2}(d) demonstrate that, under the CSPCL mechanism, intra-class content queries cluster together while maintaining necessary distances, whereas inter-class content queries repel each other.
Additionally, the content queries of the same category show a clear correlation with the category prototypes perceived by the model, i.e., the classifier weights. This indicates that our method successfully aligns the content queries with the inherent feature distribution of contraband, thereby addressing the deficiency in Deformable DETR-like models.

\begin{figure}
  \centering 
  \begin{subfigure}{2.3in}
    \includegraphics[width=\linewidth]{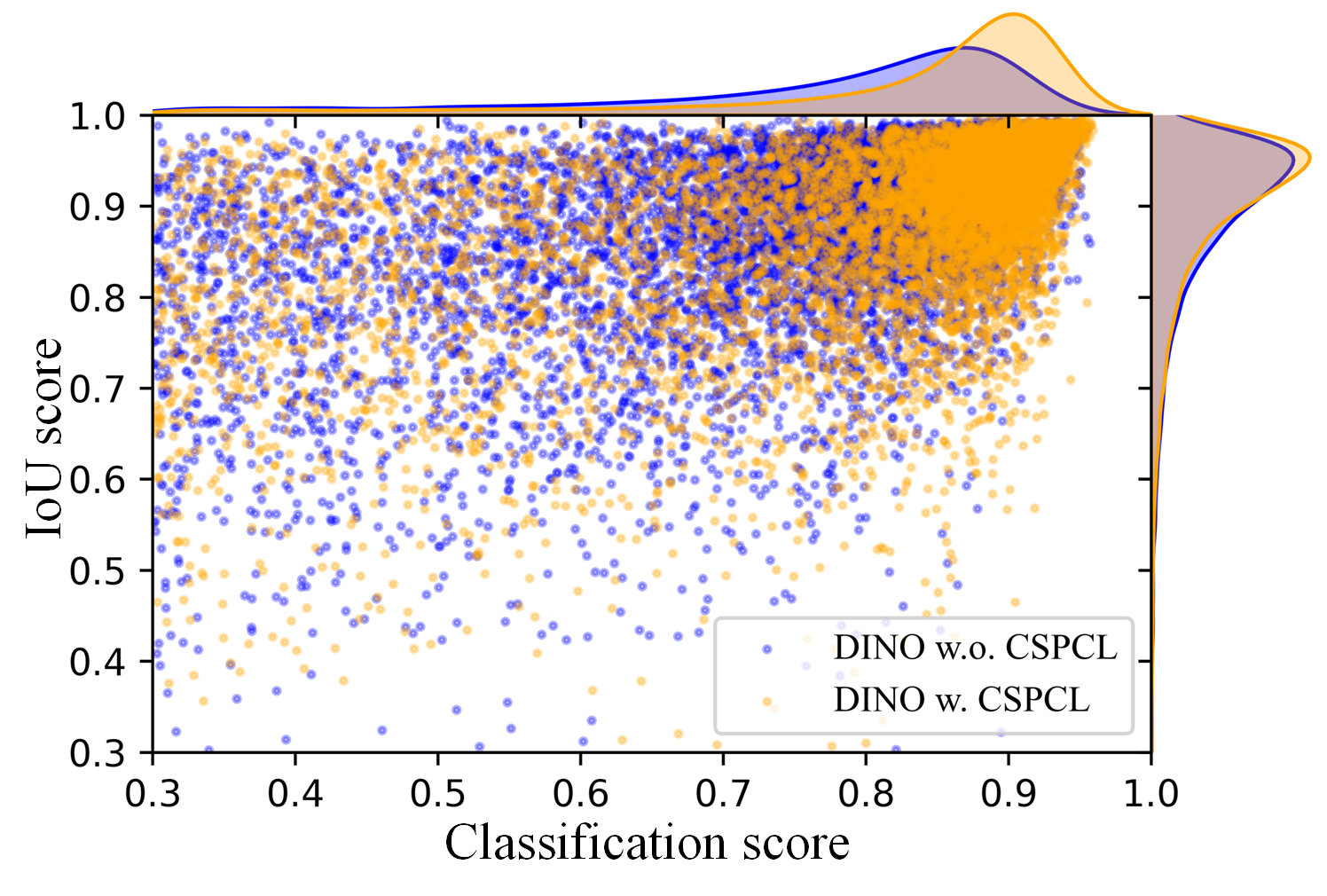}
    \caption{} 
    \label{figure Joint distribution plot}
  \end{subfigure}
  \hfill
  \begin{subfigure}{3.0in}
    \includegraphics[width=\linewidth]{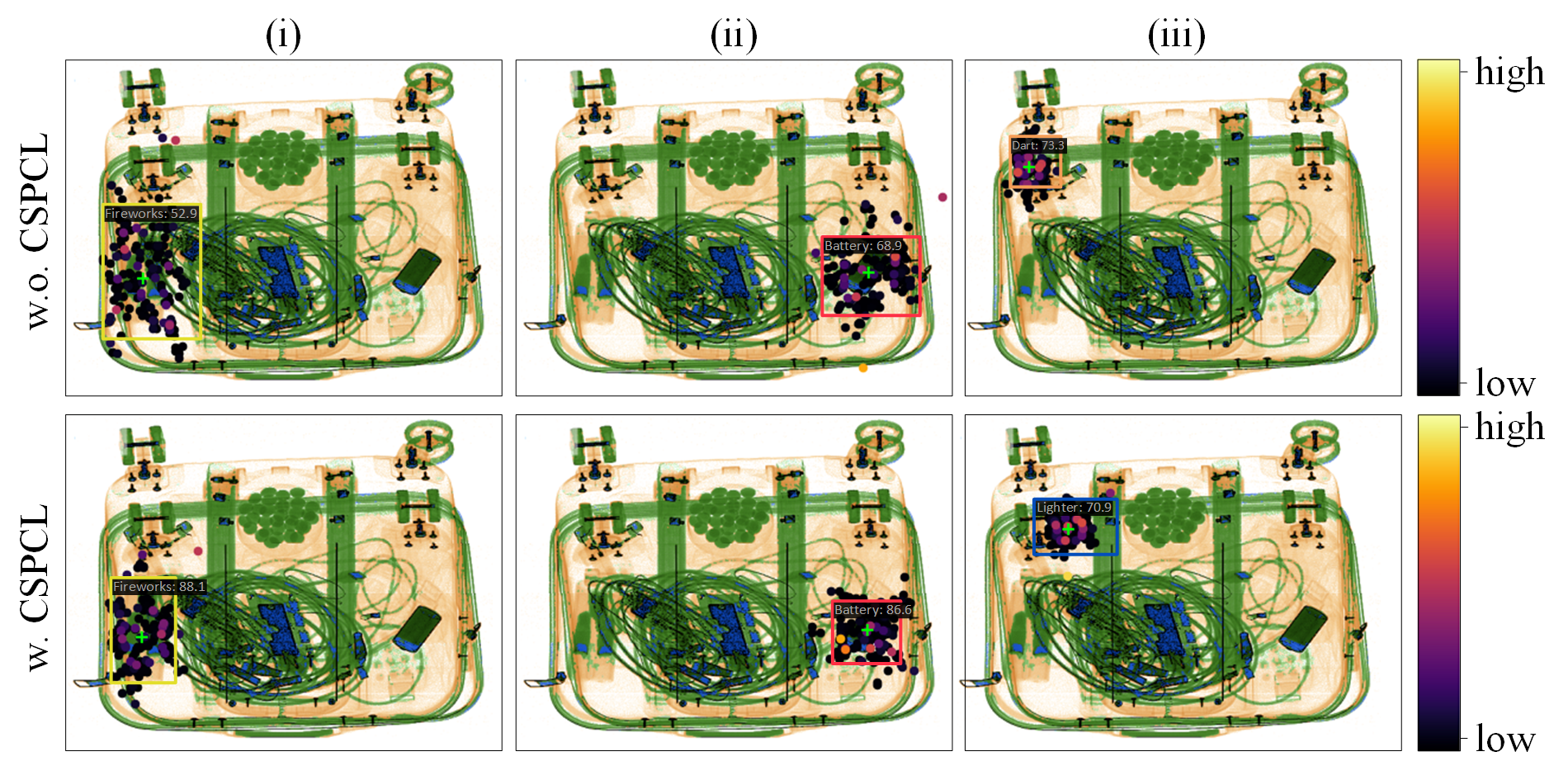}
    \caption{} 
    \label{sampling points}
  \end{subfigure}
  \caption{
  (a) The scatter plot and Kernel Density Estimation (KDE) joint distribution plot of prediction results of the final decoder layer.
\textcolor{bluejoint}{Blue} and \textcolor{orangejoint}{Orange} represent the results of DINO and DINO with the CSPCL mechanism, respectively.
(b) Visualization of deformable attention sampling points, reference points, and prediction results for corresponding group content queries in the last decoder layer. (i)–(iii) represent different groups. Each sampling point is shown as a filled circle, with color indicating its attention weight, and the reference point is marked by a green cross.
}     
\vspace{-10pt} 
\end{figure}

To analyze the effectiveness of the CSPCL mechanism for detection results, we visualize the IoU scores and classification scores of prediction results of DINO on the PIXray dataset, as shown in~\Cref{figure Joint distribution plot}.
We draw the scatter plot of prediction results whose classification scores and IoU scores are higher than 0.3, along with the Kernel Density Estimation (KDE) curves. Blue and orange represent the results of DINO and DINO with the CSPCL mechanism, respectively.
The orange points are more concentrated and distributed further to the right compared to the blue points, indicating that under the CSPCL mechanism, the content queries capture more classification semantic prior information without losing localization features, improving the accuracy of classification results. This demonstrates that our CSPCL mechanism can help the model more effectively perceive and extract foreground information of specific classes from the overlapping features in X-ray images.



\vspace{-10pt}

\section{Experiments}
\subsection{Experimental setup}\label{section Experimental setup}
\textbf{Datasets and evaluation metrics.}
    We conduct extensive experiments on four large-scale publicly available X-ray prohibited item datasets, OPIXray~\cite{OPIXray}, PIXray~\cite{PIXray}, CLCXray~\cite{CLCXray}, and PIDray~\cite{PIDray}, which are widely used in the field of prohibited item detection.
The PIXray, CLCXray, and PIDray datasets adopt the COCO~\cite{COCO} evaluation metric, AP, which measures the detector's average precision across various IoU thresholds, providing a comprehensive assessment of overall detection performance. The OPIXray dataset adopts the VOC~\cite{VOC} evaluation metric, mAP, which is calculated as the mean average precision across all categories at an IoU threshold of 0.5.

\textbf{Implementation details.}
For fair, all training and testing are conducted on the same computer platform, equipped with an NVIDIA GeForce RTX 4090 GPU, an Intel Core i9-13900K CPU, 64GB of memory, running Windows 10, and using PyTorch 1.13.1.
Furthermore, we utilize pre-trained ResNet-50 models from the official MMDetection~\cite{MMDetection} repository as the backbone, and all models are trained by the SGD optimizer with a learning rate of 0.01, a momentum of 0.9, and a weight decay of 0.1. 
In addition, all models are trained for 12 epochs, following the original training strategy, with an image size of $320 \times 320$ unless otherwise stated.

\subsection{Generalization}

\begin{wraptable}{r}{8cm}
        \vspace{-12pt} 
    \caption{Comparing classic contrastive losses with our CSP loss under CSPCL mechanism.} 
    \vspace{-5pt}
        \resizebox{1\linewidth}{!}{
    \Large
    \begin{tabular}{l|cccccc}
    \toprule
         Method&  AP & AP$_{50}$ & AP$_{75}$ & AP$_S$ & AP$_M$ & AP$_L$ \\
         \midrule
         DINO&  64.3  &86.5 &71.0 &19.3 &48.9 &73.9 \\
         DNIO + N-pair &  66.0 \bl{(+1.7)} &88.1 &72.2 &23.7 &51.3 &75.4 \\
         DINO + InfoNCE &  65.4 \bl{(+1.1)} &87.4 &72.3 &22.1 &51.1 &74.6\\
         DINO + CSP (our) & \textbf{67.5} \textbf{\bl{(+3.2)}} &\textbf{88.7} &\textbf{74.6} &\textbf{23.3} &\textbf{52.3} &\textbf{76.8}\\
         \bottomrule
    \end{tabular}
    }       
    \vspace{-8pt} 
    \label{Comparison with other contrastive losses in our CSPCL mechanism}
\end{wraptable}
\textbf{Contrastive loss.} To verify the generalization of the CSPCL mechanism for contrastive loss, we replace the proposed CSP loss with classic contrastive losses, such as N-pair loss and InfoNCE loss, to align content queries and prototypes of the same category, and evaluate them on DINO. As shown in~\Cref{Comparison with other contrastive losses in our CSPCL mechanism}, these losses still improve DINO's AP score on the PIXray dataset by $1.7\%$ and $1.1\%$, respectively, but are less effective than the $3.2\%$ improvement achieved by CSP loss. From these results, we draw two conclusions: (1) The strategy of aligning content queries with category prototypes in the CSPCL mechanism is generally effective for prohibited item detection in X-ray images; (2) Thanks to gradient truncation and the adaptive repulsion mechanism, CSP loss is better suited for the multi-class multi-sample aligning task compared to the aforementioned classic contrastive losses. More theoretical comparison analysis with other contrastive losses can be found in~\cref{appendix contrastive losses}.

\textbf{Models and datasets.} 
To verify the generalization of the CSPCL mechanism for models and datasets, we apply it to multiple representative Deformable DETR variants, including the basic Deformable DETR, the powerful DINO, RT-DETR, and the contraband-specialized AO-DETR on two prohibited item datasets PIXray and OPIXray. As shown in~\Cref{Comparison with state-of-the-art general detectors on PIXray and OPIXray}, the CSPCL mechanism improves the detection accuracy of Deformable DETR, RT-DETR, DINO, and AO-DETR on the PIXray dataset by $5.2\%$, $0.4\%$, $3.2\%$, and $0.8\%$ AP, respectively. 
Notably, CSPCL does not impose additional burdens on model complexity such as GFLOPs and PARAMs, nor does it reduce the model inference speed (FPS).
Similarly, CSPCL enhances the detection accuracy of the four models on the OPIXray dataset by $3.4\%$, $0.9\%$, $0.9\%$, and $0.8\%$ mAP, respectively.
These experimental results demonstrate that the CSPCL mechanism can effectively improve the Deformable DETR-based models' ability to handle overlapping detection tasks by aligning class prototypes and content queries.

\textbf{Comparison with SOTA Models.}
To demonstrate the enhancement effect of CSPCL on the prohibited item detection model, we chose AO-DETR as the baseline and train it with CSPCL under the condition of $512\times 512$ image size, $14$ epochs, $240$ queries to challenge other SOTA models, including general detectors and prohibited item detectors, on the largest known dataset for the prohibited item detection dataset PIDray~\cite{PIDray}. As shown in~\Cref{experiment Comparison with State-of-the-Art Methods results on PIDray}, $\mathcal{C}$-AO-DETR (AO-DETR + CSPCL) obtains the highest AP over easy, hard, and hidden sets, while outperforms in terms of Image Size and FLOPs. 
The validation on the fine-grained dataset for cutters and liquid containers, CLCXray~\cite{CLCXray}, is provided in~\cref{appendix Ablation Comparison with SOTA Models of CLCXray}.

\begin{table}
    \centering
    \caption{Generalization analysis of CSPCL mechanism on PIXray and OPIXray.}
        \resizebox{\linewidth}{!}{
    \Large
    \begin{tabular}{l|c|ccc|ccc|cccccc}
    \toprule
    \multirow{2}*{Method}&\multirow{2}*{CSPCL}&  \multirow{2}*{FPS}&\multirow{2}*{PARAMs} & \multirow{2}*{GFLOPs}&  \multicolumn{3}{c|}{PIXray}&  \multicolumn{6}{c}{OPIXray} \\
    \cmidrule{6-14}
         ~&~& ~ & ~ & ~&  AP & AP$_{50}$ & AP$_{75}$&  mAP &FO&ST&SC&UT&MU\\
         \midrule
         \multirow{2}*{Deformable DETR}&\ding{55}&60 & 52.14M & 13.47&36.6&66.4&37.1&65.6&68.1&30.1&88.9&66.0&75.0 \\
         ~&\ding{51}&60 & 52.14M & 13.47&41.8 \bl{(+5.2)}&72.3&44.5&69.0 \bl{(+3.4)}&78.4&38.8&86.0&62.8&79.1\\
         \midrule
         \multirow{2}*{RT-DETR}&\ding{55}&64 & 42.81M & 17.07&61.4&84.0&68.5&69.2&75.6&30.5&88.4&66.7&84.9\\
        ~&\ding{51}&64 & 42.81M & 17.07&61.8 \bl{(+0.4)}&84.3&68.7&70.1 \bl{(+0.9)}&76.0&34.4&88.6&67.4&84.3\\
        \midrule
         \multirow{2}*{DINO}&\ding{55}&54 & 58.38M & 26.89&64.3&86.5&71.0&77.0&81.7&56.3&90.0&71.4&85.5\\
        ~&\ding{51}&54 & 58.38M & 26.89&67.5 \bl{(+3.2)}&88.7&74.6&77.9 \bl{(+0.9)}&82.8&56.0&89.9&74.2&86.7\\
        \midrule
         \multirow{2}*{AO-DETR}&\ding{55}&54 & 58.38M & 26.89&65.6&86.1&72.0&77.7 &82.8 &57.6 &89.8 &71.2 &86.9\\
        ~&\ding{51}&54 & 58.38M & 26.89&66.4 \bl{(+0.8)}&86.8&73.6&78.5 \bl{(+0.8)} &84.3& 57.7 &90.2 &73.2& 87.1\\
         \bottomrule
    \end{tabular}
    }
        \vspace{-5pt}
    \label{Comparison with state-of-the-art general detectors on PIXray and OPIXray}
\end{table}

\begin{table}
    \centering
    \caption{Comparison with SOTA models on PIDray.}
        \resizebox{\linewidth}{!}{
    \begin{tabular}{l|c|c|cccc|cc}
    \toprule
    \multirow{2}*{Method}&\multirow{2}*{Backbone}& \multirow{2}*{Image Size} & \multicolumn{4}{c|}{AP} & \multirow{2}*{PARAMs}& \multirow{2}*{FLOPs}\\
    \cmidrule{4-7}
         ~&~& ~ & easy & hard & hidden  & overall & ~ & ~\\
         \midrule
         Faster R-CNN~\cite{Faster} & ResNeXt-101-32x4d & $1333\times800$ & 70.0 & 64.5 & 49.7 & 61.4 & 60.2M & 300.5G\\
         Mask R-CNN\cite{Mask} & ResNeXt-101-32x4d & $1333\times800$ & 77.7 & 70.0 & 53.7 & 67.0 & 101.9M & 514.7G\\
         \midrule
         FSAF~\cite{FSAF} & ResNeXt-101-64x4d & $1333\times800$ & 71.0 & 61.9 & 50.7 & 61.2 & 94.0M & 461.6G\\
         FCOS~\cite{FCOS} & ResNeXt-101-64x4d & $1333\times800$ & 72.9 & 63.4 & 51.1 & 62.5 & 89.6M & 457.0G\\
         ATSS~\cite{ATSS} & ResNet-101 & $1333\times800$ & 71.0 & 64.5 & 55.4 & 63.6 & 51.1M & 295.5G\\
         VFNet~\cite{Varifocalnet} & ResNet-101 & - & 70.3 & 62.8 & 48.6 & 60.6 & 32.74M & -\\
         TOOD~\cite{TOOD} & ResNet-101 & - & 68.5 & 63.8 & 48.9 & 60.4 & 32.04M & -\\
         \midrule
         UniFormer~\cite{UniFormer} & UniFormer-B & $1333\times800$ & 70.4 & 63.9 & 40.7 & 58.3 & 69.0M & 399.0G\\
         Deformable DETR~\cite{Deformable-DETR} & ResNet-50 & $1333\times800$ & 72.1 & 65.7 & 46.2 & 61.3 & 41.0M & 210.2G\\
         DN-DETR~\cite{DN-DETR}& ResNet-50 & $1333\times800$ &76.8 &71.0& 55.6 &67.8&47.0M& 860.0G\\
         \midrule
         SDANet~\cite{PIDray} & ResNet-101 & $500\times500$ & 71.2 & 64.2 & 49.5 & 61.6 & - & -\\
         ForkNet~\cite{ForkNet}& ResNet-101 & $500\times500$ & 75.0 & 66.9 & 58.6 & 66.8 & - & -\\
         AO-DETR~\cite{AO-DETR} & Swin-L & $512\times512$ & 81.8& 72.6 &55.7 &70.0& 229.0M& 276.5G\\
        $\mathcal{C}$-AO-DETR (ours) & Swin-L & $512\times512$ & \textbf{82.2} &\textbf{73.3} &\textbf{ 56.9} &\textbf{70.8}  & 229.0M & 276.5G\\
         \bottomrule
    \end{tabular}
    }
        \vspace{-10pt}
    \label{experiment Comparison with State-of-the-Art Methods results on PIDray}
\end{table}
\subsection{Ablation experiments}
\label{appendix Ablation experiments}
To obtain the optimal hyperparameters of CSP loss, we conduct extensive ablation experiments about integrating CSPCL into DINO on the PIXray dataset. As shown in~\Cref{Ablation study of ITA}, we first independently insert ITA loss at the $0$-th decoder layer to adjust the truncation factor $\gamma$, with the optimal truncation factor $\gamma^*= 5\times10^{-3}$, achieving $65.4\%$ AP on the PIXray dataset. Similarly, as shown in~\Cref{Ablation study of IAR}, the independent insertion of IAR loss yields the optimal temperature factor $\tau^*=0.3$, achieving $65.6\%$ AP. Furthermore,~\Cref{Ablation study of ITA and IAR results} shows that when both ITA loss and IAR loss are used simultaneously, i.e., the full version of CSP loss, the model's AP increases by $2.4\%$, which is higher than the individual use of ITA loss and IAR loss. This demonstrates the compatibility of the two loss functions.
Subsequently, we adjust the target layer set where the CSPCL is inserted. As shown in~\Cref{Ablation experiments of target layer set}, the model achieves $67.2\%$ AP by aligning the content queries of all decoder layers with the category prototypes better than aligning any single layer. Notably, when aligning only one layer, shallower layers, such as layer $0$, tend to perform better. This could be because deeper layers, such as layer $5$, involve content queries that more directly contribute to detection results, and indiscriminate alignment might disrupt the spatial features perceived by the model.
Finally, as shown in~\Cref{Ablation experiments for ITA alpha} and~\Cref{Ablation experiments for IAR beta}, we adjust the parameters $\alpha$ and $\beta$ in~\cref{equation CSPCL} to control the contribution ratio of the two losses. The optimal values for $\alpha^*$ and $\beta^*$ are $1$ and $0.5$, respectively. Overall, the CSPCL mechanism with the optimal hyperparameters can help DINO achieve the highest AP score of $67.5\%$ on PIXray.
\begin{table*}[h]
    \caption{Ablation study about the CSPCL mechanism on DINO~\cite{DINO} on the PIXray~\cite{PIXray} dataset. $L$ means all decoder layers. The superscript `$*$' represents the optimal value of the hyper-parameter.}
    \centering
    \resizebox{1\linewidth}{!}{

      \begin{minipage}[t]{0.38\linewidth}
    \centering
    \resizebox{\linewidth}{!}{
        \begin{tabular}{c|ccc}
            \toprule
            $\gamma$  & AP & AP$_{50}$ & AP$_{75}$ \\
            \midrule
            $5\times 10^{-2}$ &64.9 \bl{(+0.6)} &86.1 &72.6 \\
            $\mathbf{5\times 10^{-3}}$ &\textbf{65.4} \textbf{\bl{(+1.1)}} &\textbf{86.9} &\textbf{72.8}  \\ 
            $5\times 10^{-4}$ &65.2 \bl{(+0.9)}&86.6&72.6 \\ 
            $5\times 10^{-5}$ &64.8 \bl{(+0.5)} &86.3 &72.4 \\
            \bottomrule
        \end{tabular}
        }
        \subcaption{Ablation study of ITA.}
        \label{Ablation study of ITA}

\end{minipage}%
    \begin{minipage}[t]{0.33\linewidth}
        \resizebox{1\linewidth}{!}{
    \begin{tabular}{c|ccc}
    \toprule
     $\tau$  & AP & AP$_{50}$ & AP$_{75}$\\
    \midrule
    0.1&64.6 \bl{(+0.3)}&85.5&72.2\\
   \textbf{ 0.3}&\textbf{65.6} \textbf{\bl{(+1.3)}} &\textbf{86.8} &\textbf{73.3}\\
    1& 65.3 \bl{(+1.0)}&86.1 &73.1\\
    3& 64.5 \bl{(+0.2)}& 86.1 &71.7\\
    \bottomrule
    \end{tabular}
    }    
    \subcaption{Ablation study of IAR.}
    \label{Ablation study of IAR}
    \end{minipage}
        \begin{minipage}[t]{0.36\linewidth}

    \centering
    \resizebox{1\linewidth}{!}{
    \Large
    \begin{tabular}{cc|ccc}
    \toprule
     ITA & IAR & AP & AP$_{50}$ & AP$_{75}$\\
    \midrule
    \ding{55}&\ding{55}&64.3 & 86.5 & 71.0\\
    \ding{51}&\ding{55}&65.4  \bl{(+0.9)} & 86.9 & 72.8\\
    \ding{55}&\ding{51}&65.6  \bl{(+1.1)} & 86.8 & 73.3\\
    \ding{51}&\ding{51}&\textbf{66.7} \textbf{\bl{(+2.4)}} & \textbf{87.5} & \textbf{74.4}\\
    \bottomrule
    \end{tabular}
    }
    \subcaption{Ablation study of ITA and IAR.}
    \label{Ablation study of ITA and IAR results}
    \end{minipage}
    }
    \resizebox{1\linewidth}{!}{
    
        \begin{minipage}[t]{0.39\linewidth}
        \centering
    \begin{tabular}{c|ccc}
    \toprule
     $\hat{L}$  & AP & AP$_{50}$ & AP$_{75}$\\
    \midrule
     
    0&66.7 \bl{(+2.4)} &87.5 &74.4\\
    1&66.0 \bl{(+1.7)} &87.0 &73.5\\
    5&60.6 \bl{(-3.7)} &82.9 &67.0\\
    $\mathbf{L}$&\textbf{67.2} \textbf{\bl{(+2.9)}} &\textbf{88.4} &\textbf{74.4}\\
    \bottomrule
    \end{tabular}
    \subcaption{Ablation study of the target layer set.}
    \label{Ablation experiments of target layer set}
    \end{minipage}
      \begin{minipage}[t]{0.35\linewidth}
    \centering

        \begin{tabular}{c|ccc}
            \toprule
             $\alpha$  & AP & AP$_{50}$ & AP$_{75}$ \\
            \midrule
             5 &66.6 \bl{(+2.3)} &88.1 &73.7 \\
             \textbf{1}   &\textbf{67.2} \textbf{\bl{(+2.9)}}& \textbf{88.4} &\textbf{74.4}\\
             0.5 &66.7 \bl{(+2.4)} &87.5 &74.4 \\
             0.1   & 65.1 \bl{(+0.8)} &86.8 &72.3 \\
            \bottomrule
        \end{tabular}
        \subcaption{$\beta= 1$ , $\gamma^* = 5\times10^{-3}$, $\tau^* = 0.3$}

        \label{Ablation experiments for ITA alpha}
    \end{minipage}
    \hspace{10pt}
    \begin{minipage}[t]{0.36\linewidth}
        \centering
        \begin{tabular}{c|ccc}
            \toprule
            $\beta$ & AP & AP$_{50}$ & AP$_{75}$ \\
            \midrule
            5 &66.3 \bl{(+2.0)} &87.6 &73.3 \\
            1 &67.2 \bl{(+2.9)}& 88.4 &74.4 \\
            \textbf{0.5} &\textbf{67.5} \textbf{\bl{(+3.2)}} &\textbf{88.7} &\textbf{74.6}  \\
            0.1   & 66.7 \bl{(+2.4)} &87.9 &73.8 \\
            \bottomrule
        \end{tabular}
        \subcaption{$\alpha^*=1$, $\gamma^*=5\times 10^{-3}$, $\tau^* = 0.3$}
        \label{Ablation experiments for IAR beta}
    \end{minipage}
}
\vspace{-10pt} 
\end{table*}
\subsection{Visualization}
To investigate the effect of the CSPCL mechanism on how content queries perceive and detect foreground prohibited items, we visualize the sampling points, reference points, and detection results in the last decoder layer for both DINO and ``DINO+CSPCL'' models, as shown in~\cref{sampling points}. 
Specifically, columns (i) and (ii) demonstrate that, compared to the original DINO model, the sampling points responsible for feature extraction in the ``DINO+CSPCL'' are more concentrated on the contraband objects themselves, and the confidence and localization accuracy of the detection results are higher. In column (iii), under the condition of severe background overlap, DINO mistakenly identifies a ``metal shaft'' in the background as a ``dart'', while the ``DINO+CSPCL'' focuses on the less conspicuous ``lighter''. These results indicate that the CSPCL mechanism helps the model focus on foreground information and enhances its ability to the anti-overlapping detection task.

\section{Conclusion}
In this paper, for the prohibited item detection task based on X-ray images, we propose a plug-and-play CSPCL mechanism that aligns the classifier weights and content queries in Deformable DETR-based models. This mechanism supplements and clarifies the semantic information responsible for classification within the content queries, thereby improving the model's ability to perceive foreground information in overlapping features.
Our proposed CSP loss outperforms classic contrastive losses and is better suited for the multi-class multi-sample alignment task.
The CSPCL mechanism has strong generalizability, and it can enhance the anti-overlapping detection ability of various models on four datasets by classic contrastive losses, all without increasing model complexity.

\section*{Acknowledgments}
This work is supported by the National Natural Science Foundation of China under Grant U22A2063, 62173083, 62276186, and 62206043; the China Postdoctoral Science Foundation under No.2023M730517 and 2024T170114; the Liaoning Provincial ``Selecting the Best Candidates by Opening Competition Mechanism" Science and Technology Program under Grant 2023JH1/10400045; the Fundamental Research Funds for the Central Universities under Grant N2424022; the Major Program of National Natural Science Foundation of China (71790614) and the 111 Project (B16009); and the scholarship of China Scholarship Council (202506080096).
\bibliographystyle{IEEEtran}
\bibliography{main}




\appendix
\newpage
\section*{Appendix}
\section{Comparison with SOTA Models of CLCXray}
\label{appendix Ablation Comparison with SOTA Models of CLCXray}
We further conduct experiments to verify the superiority of $\mathcal{C}$-AO-DETR on the fine-grained dataset for cutters and liquid containers, CLCXray~\cite{CLCXray}.
As shown in~\cref{experiment Comparison with State-of-the-Art Methods results on CLCXray}, $\mathcal{C}$-AO-DETR obtains the highest AP of $63.1\% $ among general objectors, including models like Cascade R-CNN, Dynamic R-CNN, and VarifocalNet, as well as prohibited item detectors, such as DOAM, LAreg, and ForkNet.
It is worth mentioning that, although our model uses Swin-L as the backbone, resulting in a relatively high number of parameters, the smaller input image size of 512×512 reduces the model's computational complexity, making it slightly lower than the previously state-of-the-art dual-backbone prohibited item detector, ForkNet.
\begin{table*}[h]
    \caption{Comparison with state-of-the-art general detectors on CLCXray.  }
    \centering
    \resizebox{1\linewidth}{!}{
    \begin{tabular}{l|c|cccccc|cc}
    \toprule
    Method & Backbone & AP & AP$_{50}$ & AP$_{75}$ & AP$_S$ & AP$_M$ & AP$_L$ & PARAMs & FLOPs \\
    \midrule
    Cascade R-CNN~\cite{Cascade}& ResNet-50& 58.4&71.4&68.4&6.8&31.5&63.8&77.06M&1789.9G \\
    Dynamic R-CNN~\cite{Dynamic}&ResNet-50&56.7&70.9&66.9&2.7&28.5&62.8& 41.75M&176.1G\\
    \midrule
    VarifocalNet~\cite{Varifocalnet}&ResNet-50&55.3&68.3&64.6&3.4&29.4&60.9&32.74M&159.7G\\
    TOOD~\cite{TOOD}&ResNet-50&57.9&71.1&66.5&6.7&34.8&63.4&32.01M&165.9G\\
    \midrule
    DOAM~\cite{OPIXray}&ResNet-50&54.3&68.5&63.5&31.2&27.3&59.9&-&-\\
    LAreg~\cite{CLCXray}&ResNet-50&58.5&70.9&67.7&12.6&30.5&63.8&-&-\\
    LAcls~\cite{CLCXray}&ResNet-50&59.3&71.8&68.2&23.0&32.4&64.5&-&-\\
    PIXDet~\cite{PIXDet-TIM}&CFB&60.4&72.4&69.4&25.3&31.7&66.0&-&-\\
    ForkNet~\cite{ForkNet}&R50\&R18&61.0&73.5&69.4&31.1&36.3&63.4&39.81M&224.6G\\
    ForkNet~\cite{ForkNet}&R101\&R18&62.0&74.0&71.4&35.6&\textbf{41.9}&67.7&55.95M&278.5G\\
    $\mathcal{C}$-AO-DETR(ours)& Swin-L&\textbf{63.1}&\textbf{74.6}&\textbf{72.6}&\textbf{49.4}&36.7&\textbf{68.9}&229.0M & 276.5G\\
    \bottomrule
    \end{tabular}
    }
    \label{experiment Comparison with State-of-the-Art Methods results on CLCXray}
\end{table*}

\section{Comparison with other contrastive losses.}\label{appendix contrastive losses}
We take the last two representative contrastive losses as examples to analyze the advantage of our CSP loss.

\textbf{IIC loss~\cite{IIC_loss}.}
The inter-class Repulsion function of the IIC loss is based on Kullback-Leibler divergence, as follows:
\begin{equation}
   L_{sKLD} = -\frac{1}{2} \left( D_{KL}(p^l_i \| p^u_j) + D_{KL}(p^u_j \| p^l_i) \right),
\end{equation}
where $p^l$ and $p^u$ compose negative sample pairs, while $i$ and $j$ are the sample index number.

Weakness: This method is designed for comparing only two classes of samples. In the multi-class, multi-sample space, it applies the same strength of inter-class repulsion across all categories, and does not ensure that inter-class content queries of similar categories maintain sufficient differences to learn discriminative identifying features.

The intra-class Attraction function of IIC loss is as follows:
\begin{equation}
L=\frac{1}{N}\sum_{i=1}^N D_{KL}(p_l\|\hat{p_l}) + D_{KL}(\hat{p_l}\|p_l),
\end{equation}
where $\hat{p_l} $ is the probability distribution of augmented data, which forms positive sample pairs with $p_l$. This constraint helps the model learn consistent feature representations of intra-class.

Weakness: It doesn't have the gradient truncation mechanism like our ITA loss (\cref{equation intra}) to prevent excessive homogenization of intra-class content queries.

\textbf{ICE loss~\cite{ICE_loss}:}
The intra-class Attraction function of the ICE loss: 
\begin{equation}
L_{hins}=E\left[-\log\left(\frac{\exp\left(\langle f_i,m_k^i\rangle /\tau_{h_{ins}} \right)}{\sum_{j=1}^{J+1} \exp \left(\langle f_i,m_j\rangle/\tau_{h_{ins}} \right)}\right)\right],
\end{equation}
where $m_k^i$ represents the hardest positive instance for anchor $f_i$, and $\tau_{h_{ins}}$ is the temperature hyperparameter. This method is essentially the InfoNCE loss~\cite{InfoNCE} with a special construction method of positive sample pairs, and it only considers the furthest intra-class sample from the anchor/center. 

Weakness: It will also lead the intra-class content queries to lose the necessary differences, as it lacks a gradient truncation mechanism like our ITA loss, which prevents intra-class homogenization.

The intra-class Attraction function of the ICE loss:
\begin{equation}
Ls_{ins} = D_{KL}(P \| Q),
\end{equation}
where $P$ and $Q$ represent the distributions of inter-instance similarities before and after augmentation. It is based on KLD loss and its defects are the same as previously mentioned IIC loss.

Overall, compared with other existing contrastive losses, our CSP loss, through the intra-class gradient truncation and inter-class adaptive repulsion mechanisms, ensures that intra-class content queries do not become homogenized, preserving the necessary feature diversity within the class, while also ensuring that inter-class content queries align with the class prototypes' feature distribution, learning discriminative identifying features for similar categories.

\section{The significance of the alignment.}
Aligning class prototypes with content queries essentially utilizes class prototypes to attract the content queries to prevent them from drifting into irrelevant regions in the feature space specific to the corresponding category, thereby correcting and supplementing the missing semantic information. Specifically, this alignment serves two purposes as follows:
\begin{enumerate}
    \item Correcting the misguiding information caused by overlapping background features: After being trained on X-ray images with overlapping phenomena, the content queries can become ambiguous and misled by irrelevant background information. Aligning the content queries with class prototypes enables the queries to focus on the features specific to their corresponding categories, thereby correcting the noisy information in the queries caused by the overlapping background in the training phase.
    \item Supplying the missing semantic information caused by the integration of positional encoding information in the decoder phase: As shown in~\Cref{subsection explore}, in the training phase, more and more positional encoding information is integrated into the content queries, which leads to the missing and catastrophic forgetting of the original class semantic information. The class prototypes are composed of classifier weights, which have learned the reliable inherent class semantic priors. Therefore, the alignment can ensure that the content queries do not deviate too much from the category semantic priors, supplying the missing and forgetting category semantic information.

\end{enumerate}


As shown in~\Cref{figure t-SNE 2}(c), the content queries of the original DINO are scattered randomly in the feature space and are far from the class prototypes (indicating low correlation between them). 
In contrast, as shown in~\Cref{figure t-SNE 2}(d), after the alignment, the intra-class queries are clustered and distributed around their corresponding class prototypes (classifier weights). This also demonstrates that, through the alignment effect of CSPCL, the content queries are supplemented with effective category semantic information.
\section{Limitations}
The CSPCL mechanism performs exceptionally well on X-ray image datasets with overlapping instances, typically helping to improve the model's AP (Average Precision) by over 1\%. However, on natural light image datasets such as COCO~\cite{COCO}, where there is no feature coupling phenomenon, CSPCL only contributes to an AP improvement of about 0.1\%. We are currently delving into the reasons behind this and exploring ways to further refine the CSPCL strategy and CSP loss to extend its application to general object detection datasets.

\newpage
\section*{NeurIPS Paper Checklist}

\begin{enumerate}

\item {\bf Claims}
    \item[] Question: Do the main claims made in the abstract and introduction accurately reflect the paper's contributions and scope?
    \item[] Answer: \answerYes{}
    \item[] Justification: We have discussed them on page 1 and page 2.

\item {\bf Limitations}
    \item[] Question: Does the paper discuss the limitations of the work performed by the authors?
    \item[] Answer: \answerYes{}
    \item[] Justification: We have discussed about that in the supplementary material.

\item {\bf Theory assumptions and proofs}
    \item[] Question: For each theoretical result, does the paper provide the full set of assumptions and a complete (and correct) proof?
    \item[] Answer: \answerNA{}
    \item[] Justification: We do not have theoretical contributions in this work, where our contributions
are validated with experiments.

    \item {\bf Experimental result reproducibility}
    \item[] Question: Does the paper fully disclose all the information needed to reproduce the main experimental results of the paper to the extent that it affects the main claims and/or conclusions of the paper (regardless of whether the code and data are provided or not)?
    \item[] Answer: \answerYes{}
    \item[] Justification: All the hyper-parameters and network organizations are provided in the main
paper and the appendix. Additionally, the code is included in the supplementary materials

\item {\bf Open access to data and code}
    \item[] Question: Does the paper provide open access to the data and code, with sufficient instructions to faithfully reproduce the main experimental results, as described in supplemental material?
    \item[] Answer: \answerNo{}
    \item[] Justification:  The code is included in the supplementary materials, and it will be open sourced upon acceptance of the paper.


\item {\bf Experimental setting/details}
    \item[] Question: Does the paper specify all the training and test details (e.g., data splits, hyperparameters, how they were chosen, type of optimizer, etc.) necessary to understand the results?
    \item[] Answer: \answerYes{}
    \item[] Justification: The experimental setup and implement details are provided in~\cref{section Experimental setup}.

\item {\bf Experiment statistical significance}
    \item[] Question: Does the paper report error bars suitably and correctly defined or other appropriate information about the statistical significance of the experiments?
    \item[] Answer: \answerNo{}
    \item[] Justification: We follow existing related works for the setting of error bars.

\item {\bf Experiments compute resources}
    \item[] Question: For each experiment, does the paper provide sufficient information on the computer resources (type of compute workers, memory, time of execution) needed to reproduce the experiments?
    \item[] Answer: \answerYes{}
    \item[] Justification: We provide information on the compute memory and model efficiency in~\cref{section Experimental setup} and~\cref{experiment Comparison with State-of-the-Art Methods results on PIDray}.
    
\item {\bf Code of ethics}
    \item[] Question: Does the research conducted in the paper conform, in every respect, with the NeurIPS Code of Ethics \url{https://neurips.cc/public/EthicsGuidelines}?
    \item[] Answer: \answerYes{}
    \item[] Justification: We have reviewed that and claim we conform that Code of Ethics.

\item {\bf Broader impacts}
    \item[] Question: Does the paper discuss both potential positive societal impacts and negative societal impacts of the work performed?
    \item[] Answer: \answerNo{}
    \item[] Justification: Our work is only for academic research purpose.
    
\item {\bf Safeguards}
    \item[] Question: Does the paper describe safeguards that have been put in place for responsible release of data or models that have a high risk for misuse (e.g., pretrained language models, image generators, or scraped datasets)?
    \item[] Answer: \answerNA{}
    \item[] Justification: The paper does not have such risks.

\item {\bf Licenses for existing assets}
    \item[] Question: Are the creators or original owners of assets (e.g., code, data, models), used in the paper, properly credited and are the license and terms of use explicitly mentioned and properly respected?
    \item[] Answer: \answerYes{}
    \item[] Justification: We have cited them in the references.

\item {\bf New assets}
    \item[] Question: Are new assets introduced in the paper well documented and is the documentation provided alongside the assets?
    \item[] Answer: \answerNA{}
    \item[] Justification: We use and cite existing datasets in this work. Other assets including
code/model will be released after submitting.

\item {\bf Crowdsourcing and research with human subjects}
    \item[] Question: For crowdsourcing experiments and research with human subjects, does the paper include the full text of instructions given to participants and screenshots, if applicable, as well as details about compensation (if any)? 
    \item[] Answer: \answerNA{}
    \item[] Justification: This paper does not involve crowdsourcing nor research with human subjects.

\item {\bf Institutional review board (IRB) approvals or equivalent for research with human subjects}
    \item[] Question: Does the paper describe potential risks incurred by study participants, whether such risks were disclosed to the subjects, and whether Institutional Review Board (IRB) approvals (or an equivalent approval/review based on the requirements of your country or institution) were obtained?
    \item[] Answer: \answerNA{}
    \item[] Justification: This paper does not involve crowdsourcing nor research with human subjects.

\item {\bf Declaration of LLM usage}
    \item[] Question: Does the paper describe the usage of LLMs if it is an important, original, or non-standard component of the core methods in this research? Note that if the LLM is used only for writing, editing, or formatting purposes and does not impact the core methodology, scientific rigorousness, or originality of the research, declaration is not required.
    \item[] Answer: \answerNA{}
    \item[] Justification: The core method development in this research does not involve LLMs as any important, original, or non-standard components.

\end{enumerate}

\end{document}